\documentclass[journal]{IEEEtran}
\usepackage{amssymb}
\usepackage{amsmath}
\usepackage{amsfonts}
\usepackage{cite}
\usepackage{graphics}
\usepackage{epsf}
\usepackage{float}
\usepackage[pdftex]{graphicx}
\usepackage[small]{caption2}
\usepackage{bm}
\usepackage{acronym}
\usepackage{multirow}
\usepackage{color}
\usepackage[normalem]{ulem}
\usepackage{array}

\newcommand{\be}{\begin{equation}}
\newcommand{\ee}{\end{equation}}
\newcommand{\bea}{\begin{eqnarray}}
\newcommand{\eea}{\end{eqnarray}}
\newcommand{\beaa}{\begin{eqnarray*}}
\newcommand{\eeaa}{\end{eqnarray*}}

\acrodef{1D}[1D]{one-dimensional}
\acrodef{2D}[2D]{two-dimensional}
\acrodef{GTBWT}[GTBWT]{generalized tree-based wavelet transform}
\acrodef{RTBWT}[RTBWT]{redundant tree-based wavelet transform}

\title{Redundant Wavelets on Graphs and High Dimensional Data Clouds}

\begin{document}
\sloppy

\author{Idan Ram, Michael~Elad,~\IEEEmembership{Senior~Member,~IEEE}, and Israel~Cohen,~\IEEEmembership{Senior~Member,~IEEE}
\thanks{This research was supported by the Israel Science Foundation (grant no.
1130/11).

I. Ram and I. Cohen are with the Department of Electrical Engineering, Technion --
Israel Institute of Technology, Technion City, Haifa 32000, Israel. E-mail
addresses: idanram@tx.technion.ac.il (I. Ram), icohen@ee.technion.ac.il
(I. Cohen); tel.: +972-4-8294731; fax: +972-4-8295757.
M. Elad is with the Department of Computer Science, Technion --
Israel Institute of Technology, Technion City, Haifa 32000, Israel. E-mail
address: elad@cs.technion.ac.il
} }

\maketitle

\begin{abstract}

In this paper, we propose a new redundant wavelet transform applicable to scalar functions defined on high dimensional coordinates,
weighted graphs and networks.
The proposed transform utilizes the distances between the given data points.
We modify the filter-bank decomposition scheme of the redundant wavelet transform by adding in each decomposition level linear operators that reorder the approximation coefficients.
These reordering operators are derived by organizing the tree-node features so as to shorten the path that passes through these points.
We explore the use of the proposed transform to image denoising,
and show that it achieves denoising results that are close to those obtained with the BM3D algorithm.

\end{abstract}


\section{Introduction}

Signal processing problems may involve inference of an unknown
scalar target function defined on non-uniformly sampled high-dimensional grid, a graph or a network.
A major challenge in processing functions on topologically complicated coordinates,
is to find efficient methods to represent and learn them.
Let $\mathbf{X}=\left\{\mathbf{x}_1,...,\mathbf{x}_N\right\}$ be the dataset such that $\mathbf{x}_i\in \mathbb{R}^n$ are points in high-dimension,
or feature points associated with the nodes of a weighted graph or network.
Also, let $f:\mathbf{X}\rightarrow \mathbb{R}$ be a scalar function defined on the above coordinates,
and let $\mathbf{f}=[f_1,\ldots,f_N]^T$, where $f_i=f(\mathbf{x}_i)$.
A key assumption in this work is that under a distance measure $w(\cdot,\cdot)$ in $\mathbb{R}^n$,
proximity between the two coordinates $\mathbf{x}_i$ and $\mathbf{x}_j$ implies proximity between their corresponding values $f_i$ and $f_j$.
The goal in this work is to develop a redundant wavelet transform that can efficiently represent the high-dimensional function $\mathbf{f}$.
Efficiency here implies sparsity, i.e. representing $\mathbf{f}$ accurately with as few as possible wavelet coefficients.

In our previous work \cite{ram2011generalized}, we have introduced the \ac{GTBWT}, which is a critically sampled (in fact, unitary) wavelet transform applicable to functions defined on irregularly sampled grid of coordinates.
We have shown that this transform requires less coefficients than both the \ac{1D} and \ac{2D} separable wavelet transforms to represent an image,
and is useful for image denoising.
The main limitation of the \ac{GTBWT} is sensitivity to translation.
Indeed, in order to obtain a smooth denoising result in \cite{ram2011generalized}, we utilized a redundant representation obtained by applying several random variants of the \ac{GTBWT} to the noisy image.
This approach is effectively similar to applying a redundant transform to the image in a rather cumbersome and computationally intensive manner.

In this paper, we introduce a \ac{RTBWT}, which extends the redundant wavelet transform \cite{mallat2009wavelet},\cite{fowler2005redundant},\cite{holschneider1989real},\cite{shensa1992discrete},\cite{beylkin1992representation} to scalar functions defined on high dimensional data clouds, graphs and networks.
This transform is obtained by modifying an implementation of the redundant wavelet transform,
which was proposed by Shensa \cite{shensa1992discrete} and Beylkin \cite{beylkin1992representation},
similarly to the way we modified the decomposition scheme of the orthonormal transform in \cite{ram2011generalized}.
This implementation employs a filter-bank decomposition scheme, similarly to the orthonormal discrete wavelet transform.
However, in each level of this scheme none of the coefficients are discarded.
We add in each decomposition level linear operators that reorder the approximation coefficients.
These operators are data-dependent, and are obtained using tree-like structures constructed from the data points.
Each reordering operator is derived by organizing the tree-node features in the corresponding level of the tree so as to shorten the path that passes through these points.
The reordering operators increase the regularity of the permuted approximation coefficients signals,
which cause their representation with the proposed wavelet transform to be more efficient (sparse).

We explore the use of the proposed transform to image denoising,
and show that it outperforms the algorithm proposed by Elad and Aharon \cite{elad2006imageC},\cite{elad2006image}
which is based on the K-SVD algorithm \cite{aharon2006uniqueness},\cite{aharon2006k},
and achieves denoising results that are similar to those obtained with the BM3D algorithm \cite{dabov2007image}.
We also show that the \ac{RTBWT} and \ac{GTBWT} achieve similar denoising results, while the former is computationally less-demanding.

The paper is organized as follows:
In Section II, we introduce the proposed redundant tree-based wavelet transform.
In Section III, we explore the use of this transform to image denoising, and present experimental results that demonstrate its advantages.
We summarize the paper in Section IV.

\section{Redundant Tree Based Wavelet Transform}

\subsection{Decomposition and Reconstruction schemes}

We wish to develop a redundant wavelet transform that efficiently (sparsely) represents its input signal $\mathbf{f}$,
defined on a irregularly sampled grid of coordinates.
To this end, we extend the redundant wavelet transform, similarly to the way we extended the orthonormal transform in \cite{ram2011generalized}.
We note that we construct our proposed transform by modifying an implementation of the redundant wavelet transform as proposed by Shensa \cite{shensa1992discrete} and Beylkin \cite{beylkin1992representation}, and not the well known algorithme $\grave{a}$ trous \cite{mallat2009wavelet},\cite{fowler2005redundant},\cite{holschneider1989real}.
This implementation employs a filter-bank decomposition scheme, similarly to the orthonormal discrete wavelet transform.
However, in each level of this scheme all the coefficients are retained since the highpass bands do not contain decimators,
and the decimation in the lowpass bands is replaced by a split into even and odd sequences, which are further decomposed in the next decomposition level.

\begin{figure}[t]
\centering
\includegraphics[scale=0.6]{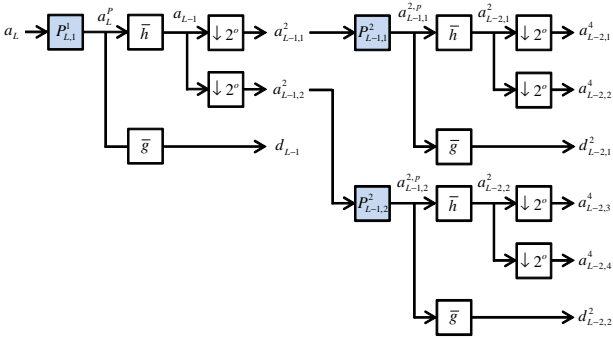}
\caption{Decomposition scheme of the proposed redundant wavelet transform.}
\label{Figure: wavelet decompositon}
\end{figure}

Fig. \ref{Figure: wavelet decompositon} describes the decomposition scheme of our proposed redundant wavelet transform.
We denote the coarsest decomposition level $\ell=1$ and the finest level $\ell=L=\text{log}_2 N + 1$.
$\mathbf{a}_\ell$ and $\mathbf{d}_\ell$ denote the approximation and detail coefficients in level $\ell$, respectively.
We start with the finest decomposition level, $\mathbf{a}_L = \mathbf{f}$, and apply the linear operator $P_{L,1}^1$,
which produces a permuted version $\mathbf{a}_L^p$ of its input vector.
$P_{\ell,s}^t$ denotes a linear operator that operates in the $s$th band, out of $t$ bands, in the $\ell$th decomposition level,
and produces a permuted version of its input vector.
These operators make the difference between our proposed wavelet decomposition scheme and the common redundant wavelet transform \cite{shensa1992discrete},\cite{beylkin1992representation}.
As we explain later, these operators "smooth" the approximation coefficients in the different levels of the decomposition scheme.
Next, we apply the wavelet decomposition filters $\bar{\mathbf{h}}$ and $\bar{\mathbf{g}}$ on $\mathbf{a}_L^p$,
and obtain the vectors $\mathbf{a}_{L-1}$ and $\mathbf{d}_{L-1}$, respectively.
Let $\downarrow 2^o$ and $\downarrow 2^e$ denote $2:1$ decimators that keep the odd and even samples of their input, respectively.
Then we employ these decimators to obtain the signals $\mathbf{a}_{L-1,1}^2$ and $\mathbf{a}_{L-1,2}^2$.
These two vectors are used as inputs for the next decomposition level.

We continue in a similar manner in the following decomposition levels.
Let $\mathbf{a}_{\ell,s}^t$ denote an approximation coefficients vector, which is found in the $s$th band (out of $t$ bands), in the $\ell$th decomposition level.
This vector is obtained by starting from the $s$th sample in $\mathbf{a}_\ell$, and keeping every $t$th sample.
Then in the $\ell$th decomposition level we decompose each of the vectors $\{\mathbf{a}_{\ell,s}^{2^{L-\ell}}\}_{s=1}^{2^{L-\ell}}$.
We first apply on each vector $\mathbf{a}_{\ell,s}^{2^{L-\ell}}$ the linear operator $P_{\ell,s}^{2^{L-\ell}}$ and obtain a permuted version $\mathbf{a}_{\ell,s}^{2^{L-\ell},p}$. We then filter $\mathbf{a}_{\ell,s}^{2^{L-\ell},p}$ with $\bar{\mathbf{h}}$ and $\bar{\mathbf{g}}$
and obtain the vectors $\mathbf{a}_{\ell-1,s}^{2^{L-\ell}}$ and $\mathbf{d}_{\ell-1,s}^{2^{L-\ell}}$, respectively.
Finally, we employ the decimators $\downarrow 2^o$ and $\downarrow 2^e$ to split each of the vectors $\{\mathbf{a}_{\ell-1,s}^{2^{L-\ell}}\}_{s=1}^{2^{L-\ell}}$ into even and odd sequences, respectively,
and obtain the set of vectors $\{\mathbf{a}_{\ell-1,s}^{2^{L-\ell+1}}\}_{s=1}^{2^{L-\ell+1}}$.

\begin{figure}[t]
\centering
\includegraphics[scale=0.6]{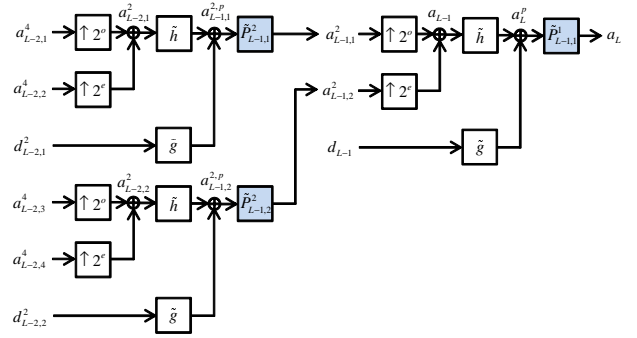}
\caption{Reconstruction scheme of the proposed redundant wavelet transform.}
\label{Figure: wavelet reconstruction}
\end{figure}

In a similar manner, Fig. \ref{Figure: wavelet reconstruction} describes the reconstruction scheme of our redundant wavelet transform.
$\tilde{\mathbf{h}}=\frac{1}{2}\mathbf{h}$ and $\tilde{\mathbf{g}}=\frac{1}{2}\mathbf{g}$, where $\mathbf{h}$ and $\mathbf{g}$ denote the wavelet reconstruction filters,
and the interpolators denoted by $\uparrow 2^o$ and $\uparrow 2^e$ place the samples of their input vector in the odd and even locations of their output vector, respectively.
Finally, the linear operator $\tilde{P}_{\ell,s}^t$ reorders a vector so as to cancel the ordering done by $P_{\ell,s}^t$,
i.e. $\tilde{P}_{\ell,s}^t=(P_{\ell,s}^t)^{-1}$.
We next describe how the linear operators $P_{\ell,s}^t$ are determined in each level of the transform.

\subsection{Building the operators $P_{\ell,s}^t$}

We wish to design the operators $P_{\ell,s}^t$ in a manner which results in an efficient (sparse) representation of the input signal by the proposed transform.
The wavelet transform is known to produce a small number of large coefficients when it is applied to piecewise regular signals \cite{mallat2009wavelet}.
Thus, we would like the operator $P_{\ell,s}^t$, applied to $\mathbf{a}_{\ell,s}^t$, to produce a signal which is as regular as possible.
We start with the finest level, and try to find the permutation that the operator $P_{L,1}^1$ applies to $\mathbf{a}_L=\mathbf{f}$.
When the signal $\mathbf{f}$ is known,
the optimal solution would be to apply a simple {\em sort} operation.
However, since we are interested in the case where $\mathbf{f}$ is not necessarily known (such as in the case where $\mathbf{f}$ is noisy, or has missing values),
we would try to find a suboptimal ordering operation, using the feature coordinates $\mathbf{x}_i$.

We recall our assumption that the distance $w(\mathbf{x}_i,\mathbf{x}_j)$ predicts the proximity between the samples $f_i$ and $f_j$.
Thus, we try to reorder the points $\mathbf{x}_i$ so that they form a smooth path,
hoping that the corresponding reordered \ac{1D} signal $\mathbf{a}_L^p$ will also be smooth.
The ``smoothness'' of the reordered signal $\mathbf{a}_L^p$ can be measured using its total variation measure
\begin{equation}
\|\mathbf{a}_L^p\|_V=\sum_{j=2}^{N} |a_L^p(j)-a_L^p(j-1)|.
\end{equation}
Let $\{\mathbf{x}_j^p\}_{j=1}^N$ denote the points $\{\mathbf{x}_i\}_{i=1}^N$ in their new order.
Then by analogy, we measure the "smoothness" of the path through the points $\mathbf{x}_j^p$ by the measure
\begin{equation}
C_{L,V}^p=\sum_{j=2}^{N} w(\mathbf{x}_j^p,\mathbf{x}_{j-1}^p).
\end{equation}
We note that in the case that $w$ is the Euclidean distance and $f$ is Lipschitz continuous,
i.e., there exists a real constant $K\geq0$ such that
\begin{equation}
|f(\mathbf{x}_i)-f(\mathbf{x}_j)|\leq K\|\mathbf{x}_i-\mathbf{x}_j\|_2
\end{equation}
for all $\mathbf{x}_i$ and $\mathbf{x}_i$ in $\mathbf{X}$, then
\begin{equation}
\sum_{j=2}^{N} |a_L^p(j)-a_L^p(j-1)|\leq K\sum_{j=2}^{N} \|\mathbf{x}_j^p-\mathbf{x}_{j-1}^p\|_2
\end{equation}
which means that $KC_{L,V}^p$ is an upper bound for $\|\mathbf{a}_L^p\|_V$.

Minimizing $C_{L,V}^p$ comes down to finding the shortest path that passes through the set of points $\mathbf{x}_j$, visiting each point only once.
This can be regarded as an instance of the traveling salesman problem \cite{cormen2001introduction},
which can become very computationally exhaustive for large sets of points.
We choose a simple approximate solution, which is to start from an arbitrary point (random or pre-defined),
and continue from each point to its nearest neighbor, not visiting any point twice.
The permutation applied by the operator $P_{L,1}^1$ is defined as the order in the found path.

In order to employ the aforementioned method to find the operators $P_{L-1,1}^2$ and $P_{L-1,2}^2$ in the $L-1$th decomposition level,
we again require feature points in order to predict the proximity between the samples of $\mathbf{a}_{L-1}$.
Since $\mathbf{a}_{L-1,1}^2$ and $\mathbf{a}_{L-1,2}^2$ are obtained from $\mathbf{a}_L$ through filtering and subsampling,
each approximation coefficient $a_{L-1}(i)$ is in fact calculated as a weighted mean of coefficients from $\mathbf{a}_L$,
where the coefficients in $\bar{\mathbf{h}}$ serve as the weights.
Thus, we calculate the feature point $\mathbf{c}_i^{L-1}$, which corresponds to $a_{L-1}(i)$, by replacing each coefficient $a_L(i)$ in this weighted mean by its corresponding feature point $\mathbf{x}_i$.
We then employ the approximate shortest path search method described above to obtain the operators $P_{L-1,1}^2$ and $P_{L-1,2}^2$ using the feature points that correspond to the coefficients in $\mathbf{a}_{L-1,1}^2$ and $\mathbf{a}_{L-1,2}^2$, respectively.

\begin{figure}[t]
\centering
\includegraphics[scale=0.3]{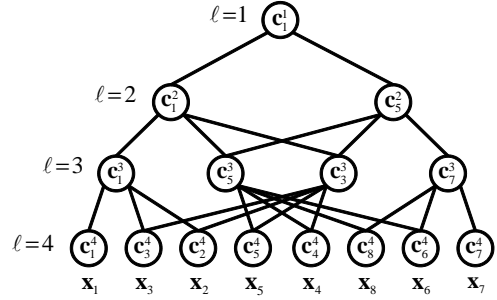}
\caption{An illustration of a ``generalized'' tree.}
\label{Figure: multipartiteGraph}
\end{figure}

We continue in a similar manner in the following decomposition levels.
In level $\ell$ we first obtain the feature points $\mathbf{c}_i^{\ell}$ as weighted means of feature points from the finer level $\ell+1$.
Then we use these feature points to obtain the operators $P_{\ell,s}^t$, running the approximate shortest path searches.
Similarly to the GTBWT decomposition scheme \cite{ram2011generalized},
the relation between the feature points in a full decomposition can be described using tree-like structures.
Each such ``generalized'' tree contains all the feature points which have participated in the calculation of a single feature point $\mathbf{c}_i^1$ from the coarsest decomposition level.
Also, each feature point in the tree level $\ell$ is connected to all the points in level $\ell+1$ that were averaged in its construction.
Fig. \ref{Figure: multipartiteGraph} shows an example of a ``generalized'' tree,
which may be obtained for a dataset of length $N=8$, using a filter $\bar{\mathbf{h}}$ of length $4$
and disregarding boundary issues in the different levels.
As the construction of these tree-like structures play an integral part in our proposed transform, we term it redundant tree-based wavelet transform (RTBWT).

We also note that the computational complexities of both the RTBWT and the GTBWT are dominated by the number of distances that need to be calculated in their wavelet decomposition schemes.
In \cite{ram2011generalized} we employed the orthonormal transforms corresponding to several randomly constructed trees in order to apply a redundant transform.
A full RTBWT decomposition, corresponding to redundancy factor of $L$, requires the calculation of $N^2-2N-\frac{1}{2}N(L-1)+1$ distances.
The method employed in \cite{ram2011generalized} requires $L\left(\frac{2}{3}N^2-N-L+\frac{4}{3}\right)$ distance calculations in order to obtain a transform with a similar redundancy factor.
Therefore for large $N$ it requires about $\frac{2}{3}L$ times more distance calculations than the RTBWT.
We next demonstrate the application of our proposed transform to image denoising.

\section{Image denoising using RTBWT}

Let $\mathbf{F}$ be an image of size $N_1\times N_2$ where $N_1 N_2=N$, and let $\tilde{\mathbf{F}}$ be its noisy version:
\begin{equation}
\tilde{\mathbf{F}}=\mathbf{F}+\mathbf{Z}.
\end{equation}
$\mathbf{Z}$ denotes an additive white Gaussian noise independent of $\mathbf{F}$ with zero mean and
variance $\sigma^2$.
Also, let $\mathbf{f}$ and $\tilde{\mathbf{f}}$ be vectors containing the pixels of $\mathbf{F}$ and $\tilde{\mathbf{F}}$, respectively, arranged in the same lexicographical order.
Our goal is to reconstruct $\mathbf{f}$ from $\tilde{\mathbf{f}}$ using the \ac{RTBWT}.
To this end, we first construct the image $\tilde{\mathbf{F}}^b$ of size $(N_1+\sqrt{n}-1)\times(N_2+\sqrt{n}-1)$ by padding $\tilde{\mathbf{F}}$ with mirror reflections of itself, and then extract the feature points $\mathbf{x}_i$ from $\tilde{\mathbf{F}}^b$.
Let $\tilde{f}_i$ be the $i$th sample in $\tilde{\mathbf{f}}$,
then we choose the point $\mathbf{x}_i$ associated with it as the $\sqrt{n} \times \sqrt{n}$ patch
around the location of $\tilde{f}_i$ in the image $\tilde{\mathbf{F}}^b$.
Next, we obtain all the operators $P_{\ell,s}^t$ employed by the RTBWT using the scheme described above.
We use the squared Euclidean distance to measure the dissimilarity between the feature points in each level,
and restrict the nearest neighbor searches performed for each patch to a surrounding square neighborhood which contains $B\times B$ patches.
This restriction decreases the computational complexity of the transform and our experiments showed that with a proper choice of $B$ it also leads to improved denoising results.

Now let $\mathbf{C}_L$ be a matrix of size $n\times N$,
containing column stacked versions of all the patches $\mathbf{x}_i$ inside the image.
In \cite{ram2011generalized} we observed that improved denoising results are obtained
by using in the denoising process all the $n$ signals located in the rows of $\mathbf{C}_L$.
These signals are the column stacked versions of the $n$ images of size $N_1\times N_2$,
whose top left pixel resides in the top left $\sqrt{n} \times \sqrt{n}$ patch in the image $\tilde{\mathbf{F}}^b$.
We apply a $9$ level decomposition to each of these subimages, and obtain $10$ coefficient matrices of size $n\times N$.
We chose to perform a $9$ level decomposition, which corresponds to a redundancy factor of $10$, over a full decomposition,
as the former is less computationally and memory demanding, but produces similar denoising results.
Next, we zero in each such matrix all the columns whose norm is smaller than a threshold T.
Note that this is different (and better performing), compared to the plain hard thresholding that is described in \cite{ram2011generalized}.
We then apply the inverse transform to each subimage, plug it into its original place in the image,
and obtain the denoised image by averaging the different values obtained for each pixel.

\begin{figure}[t]
\centering
\begin{tabular}{cc}
\includegraphics[scale=0.3]{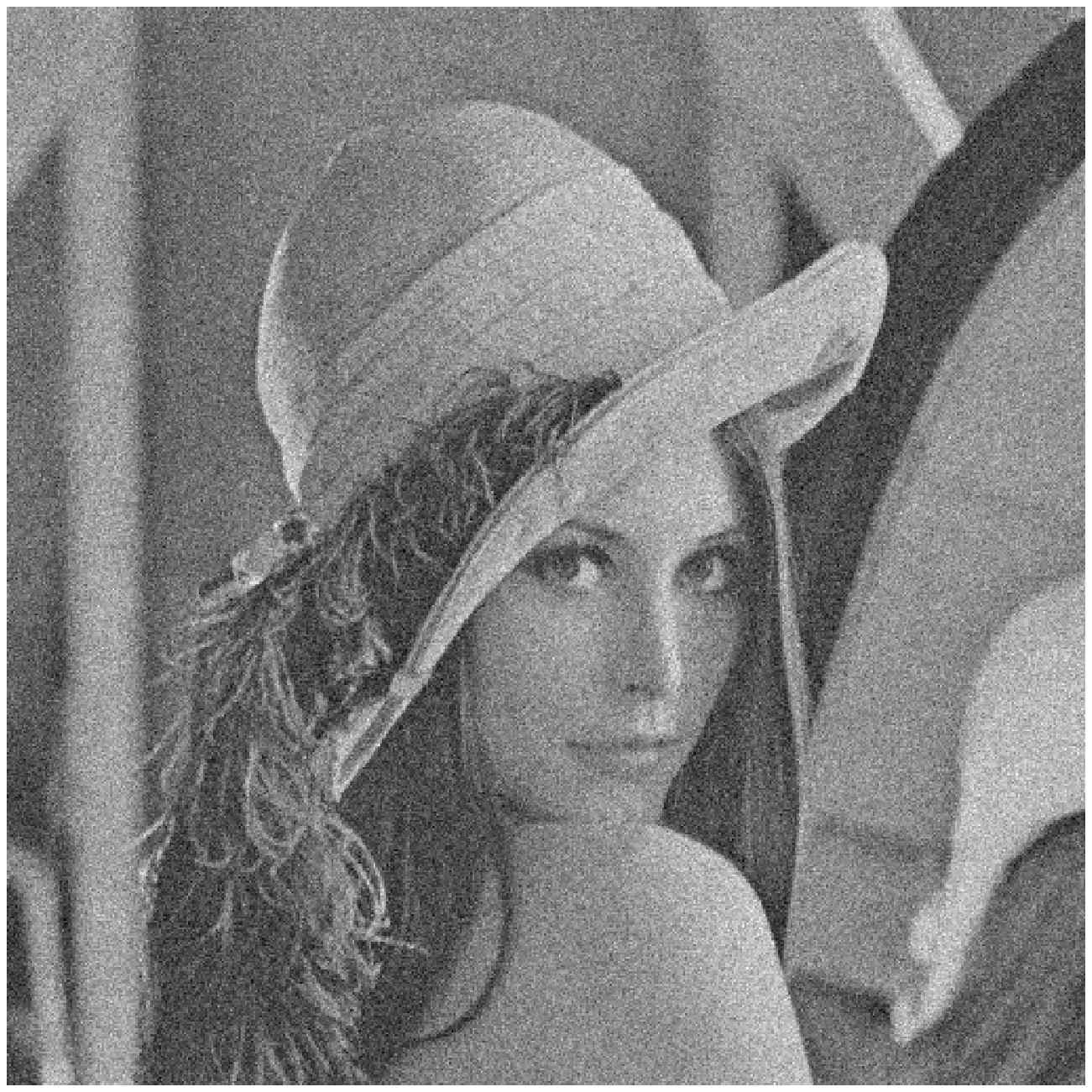} & \includegraphics[scale=0.3]{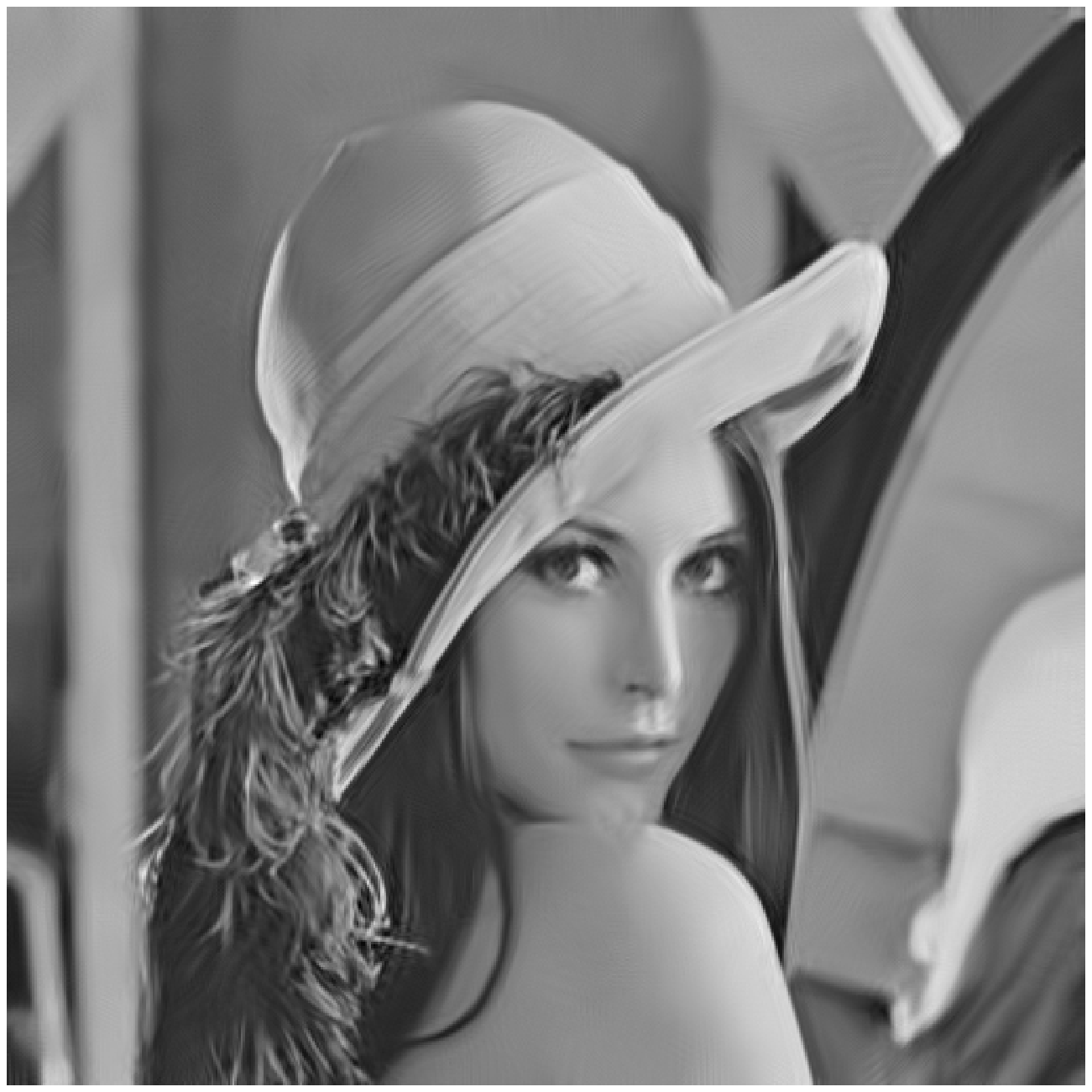}\\
{\small (a)} & {\small (b)}\\
\includegraphics[scale=0.3]{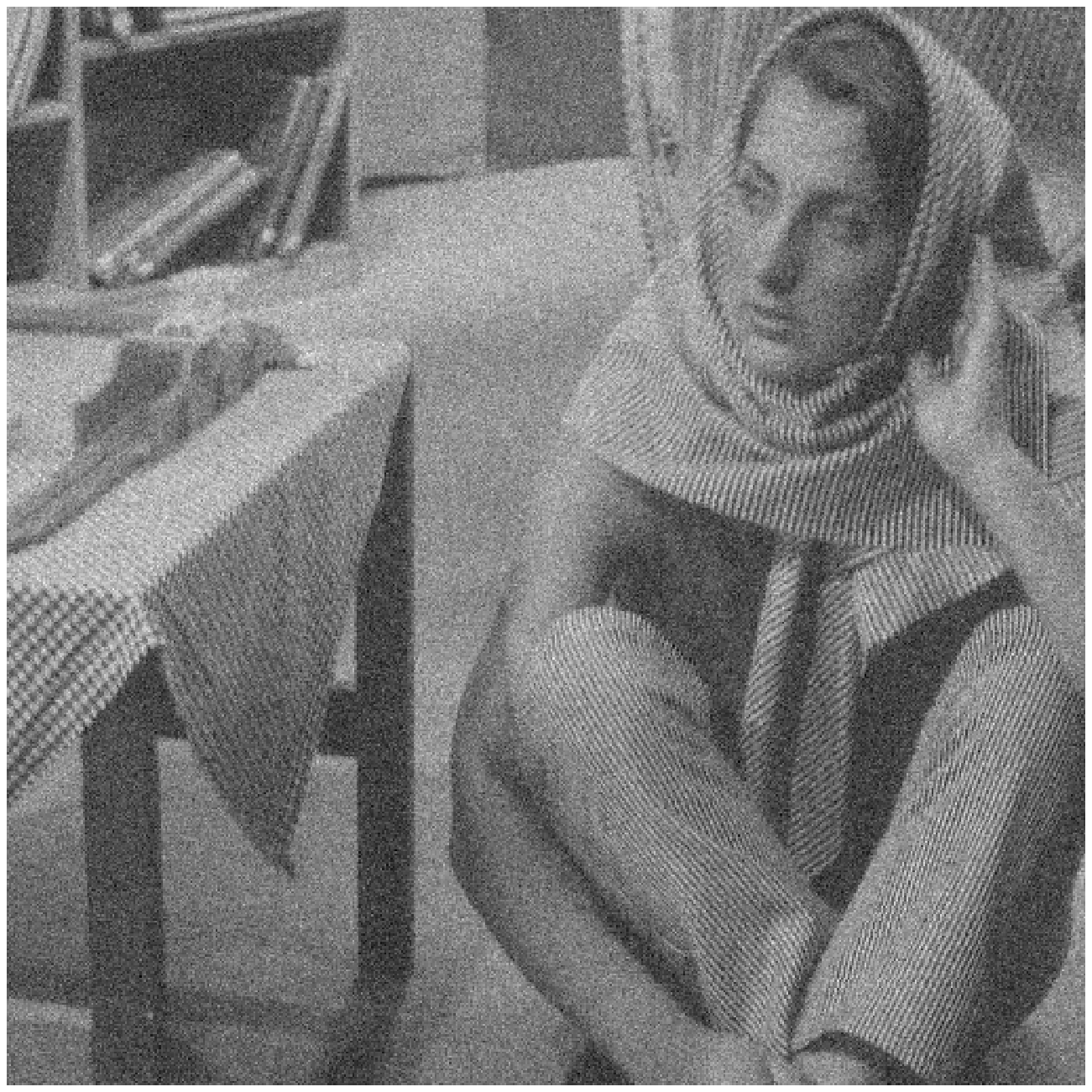} & \includegraphics[scale=0.3]{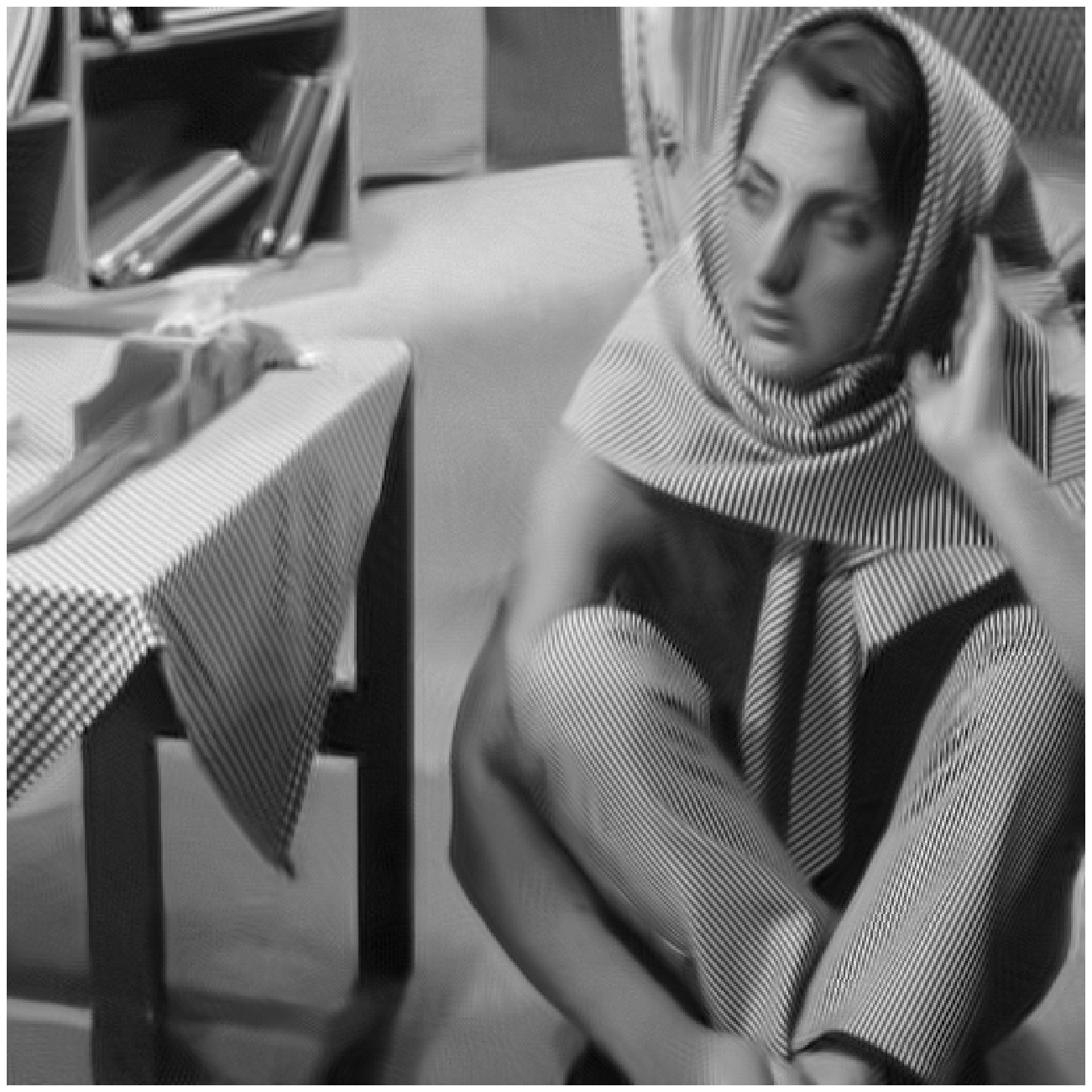}\\
{\small (c)} & {\small (d)}\\
\end{tabular}
\caption{Denoising results of noisy versions of the images Barbara and Lena ($\sigma=25$, input PSNR=20.18 dB) obtained with RTBWT with a Symmlet 8 filter:
(a) Noisy Lena. (b) Lena denoised using RTBWT (32.17 dB). (c) Noisy Barbara (20.17 dB). (d) Barbara denoised using RTBWT (30.76 dB).}
\label{Figure: full Images}
\end{figure}

In order to assess the performance of the proposed image denoising scheme
we apply it with the Symmlet 8 wavelet filter to noisy versions of the images Lena and Barbara,
with noise standard deviations $\sigma=10,25$.
The noisy and recovered images corresponding to $\sigma=25$ can be seen in Fig. \ref{Figure: full Images}.
For comparison, we also apply to the two images the algorithm proposed by Elad and Aharon \cite{elad2006imageC},\cite{elad2006image}
which is based on the K-SVD algorithm \cite{aharon2006uniqueness},\cite{aharon2006k}, the BM3D algorithm \cite{dabov2007image},
and the \ac{GTBWT} with the search neighborhood and thresholding method described above.
The PSNR of the results obtained with all the four denoising schemes are shown in Table \ref{Table: full_images_PSNR}.
It can be seen that the results obtained with both the RTBWT and the GTBWT are better than the ones obtained with the K-SVD algorithm,
and are close to the ones obtained with the BM3D algorithm.
However, the RTBWT was about $6$ times faster than the GTBWT since it required much less distance calculations.

\begin{table}[t]
\centering
\caption{Denoising Results (PSNR in dB) of Noisy Versions of
the Images Barbara and Lena Obtained with the K-SVD (top left),
BM3D (top right), GTBWT (bottom left) and RTBWT (bottom
right) Algorithms. For Each Image and Noise Level the Best Result is
Highlighted.\newline}
\begin{tabular}{|c|c|c!{\vrule width 1pt}c|c|}\hline
$\sigma$/PSNR & \multicolumn{2}{c!{\vrule width 1pt}}{Lena} & \multicolumn{2}{c|}{Barbara} \\\hline
\multirow{2}{*}{10/20.18} & 35.51  & $\mathbf{35.93}$ & 34.44  & $\mathbf{34.98}$ \\\cline{2-5}
& 35.87 & 35.88 & 34.94 & 34.93 \\\noalign{\hrule height 1pt}
\multirow{2}{*}{25/28.14} & 31.36  & 32.08 &  29.57  & 30.72 \\\cline{2-5}
& 32.16 &  $\mathbf{32.17}$ &  30.75 & $\mathbf{30.76}$ \\\hline
\end{tabular}
\label{Table: full_images_PSNR}
\end{table}

\section{Conclusion}

We have proposed a new redundant wavelet transform applicable to scalar functions defined on graphs and high dimensional data clouds.
This transform is the redundant version of the GTBWT introduced in \cite{ram2011generalized}.
We have shown that our proposed scheme can be used for image denoising,
where it achieves denoising results that are close to the state-of-the-art.
In our future work plans, we intend to:
\begin{enumerate}

\item Seek ways to improve the method that reorders the approximation coefficients in each level of the tree, replacing the proposed approximate shortest path search method.

\item Modify the denoising algorithm so it will adaptively choose the patch and search neighborhood sizes, and the denoising threshold.

\item Improve the image denoising results by using two iterations with different threshold settings.

\end{enumerate}

\bibliographystyle{IEEEtran}
\bibliography{IEEEabrv,treeWave}

\begin{thebibliography}{10}
\providecommand{\url}[1]{#1}
\csname url@samestyle\endcsname
\providecommand{\newblock}{\relax}
\providecommand{\bibinfo}[2]{#2}
\providecommand{\BIBentrySTDinterwordspacing}{\spaceskip=0pt\relax}
\providecommand{\BIBentryALTinterwordstretchfactor}{4}
\providecommand{\BIBentryALTinterwordspacing}{\spaceskip=\fontdimen2\font plus
\BIBentryALTinterwordstretchfactor\fontdimen3\font minus
  \fontdimen4\font\relax}
\providecommand{\BIBforeignlanguage}[2]{{%
\expandafter\ifx\csname l@#1\endcsname\relax
\typeout{** WARNING: IEEEtran.bst: No hyphenation pattern has been}%
\typeout{** loaded for the language `#1'. Using the pattern for}%
\typeout{** the default language instead.}%
\else
\language=\csname l@#1\endcsname
\fi
#2}}
\providecommand{\BIBdecl}{\relax}
\BIBdecl

\bibitem{ram2011generalized}
I.~Ram, M.~Elad, and I.~Cohen, ``{Generalized Tree-Based Wavelet Transform },''
  \emph{IEEE Trans. Signal Processing}, vol.~59, no.~9, pp. 4199--4209, 2011.

\bibitem{mallat2009wavelet}
S.~Mallat, \emph{A Wavelet Tour of Signal Processing, The Sparse Way}.\hskip
  1em plus 0.5em minus 0.4em\relax Academic Press, 2009.

\bibitem{fowler2005redundant}
J.~Fowler, ``The redundant discrete wavelet transform and additive noise,''
  \emph{IEEE Signal Processing Letters}, vol.~12, no.~9, pp. 629--632, 2005.

\bibitem{holschneider1989real}
M.~Holschneider, R.~Kronland-Martinet, J.~Morlet, and P.~Tchamitchian, ``A
  real-time algorithm for signal analysis with the help of the wavelet
  transform,'' in \emph{Wavelets. Time-Frequency Methods and Phase Space},
  vol.~1.\hskip 1em plus 0.5em minus 0.4em\relax Springer-Verlag, 1989, pp.
  286--297.

\bibitem{shensa1992discrete}
M.~Shensa, ``The discrete wavelet transform: Wedding the a trous and mallat
  algorithms,'' \emph{IEEE Trans. Signal Processing}, vol.~40, no.~10, pp.
  2464--2482, 1992.

\bibitem{beylkin1992representation}
G.~Beylkin, ``On the representation of operators in bases of compactly
  supported wavelets,'' \emph{SIAM J. Numer. Anal.}, vol.~29, no.~6, pp.
  1716--1740, 1992.

\bibitem{elad2006imageC}
M.~Elad and M.~Aharon, ``Image denoising via learned dictionaries and sparse
  representation,'' in \emph{IEEE Computer Society Conference on Computer
  Vision and Pattern Recognition}, vol.~1, 2006, pp. 895--900.

\bibitem{elad2006image}
------, ``Image denoising via sparse and redundant representations over learned
  dictionaries,'' \emph{IEEE Trans. Image Processing}, vol.~15, no.~12, pp.
  3736--3745, 2006.

\bibitem{aharon2006uniqueness}
M.~Aharon, M.~Elad, and A.~Bruckstein, ``On the uniqueness of overcomplete
  dictionaries, and a practical way to retrieve them,'' \emph{Linear algebra
  and its applications}, vol. 416, no.~1, pp. 48--67, 2006.

\bibitem{aharon2006k}
------, ``{K-SVD}: An algorithm for designing overcomplete dictionaries for
  sparse representation,'' \emph{IEEE Trans. Signal processing}, vol.~54,
  no.~11, p. 4311, 2006.

\bibitem{dabov2007image}
K.~Dabov, A.~Foi, V.~Katkovnik, and K.~Egiazarian, ``Image denoising by sparse
  3-d transform-domain collaborative filtering,'' \emph{IEEE Trans. Image
  Processing}, vol.~16, no.~8, pp. 2080--2095, 2007.

\bibitem{cormen2001introduction}
T.~H. Cormen, \emph{Introduction to algorithms}.\hskip 1em plus 0.5em minus
  0.4em\relax The MIT press, 2001.

\end{thebibliography}

\end{document}